\documentclass[10pt,twocolumn,letterpaper]{article}

\usepackage{iccv}
\usepackage{times}
\usepackage{epsfig}
\usepackage{graphicx}
\usepackage{amsmath}
\usepackage{amssymb}
\usepackage{kotex}
\usepackage{bbm}
\usepackage{amsmath}
\usepackage{multirow}
\usepackage{booktabs}
\DeclareMathOperator*{\argmax}{arg\,max}

\newcommand{\norm}[1]{\left\lVert#1\right\rVert}

\usepackage[pagebackref=true,breaklinks=true,letterpaper=true,colorlinks,bookmarks=false]{hyperref}

\iccvfinalcopy 

\def\iccvPaperID{****} 

\ificcvfinal\fi

\begin{document}

\title{Maximizing Cosine Similarity Between Spatial Features \\
for Unsupervised Domain Adaptation in Semantic Segmentation}

\author{
Inseop Chung\thanks{This work is conducted during the author's research internship at NAVER WEBTOON Corp. }\\
Seoul National University\\
Seoul, Korea \\
{\tt\small jis3613@snu.ac.kr}
\and
Daesik Kim\\
NAVER WEBTOON Corp.\\
Seoul, Korea\\
{\tt\small daesik.kim@webtoonscorp.com}
\and
Nojun Kwak\\
Seoul National University\\
Seoul, Korea \\
{\tt\small nojunk@snu.ac.kr}
}

\maketitle
\ificcvfinal\thispagestyle{empty}\fi

\begin{abstract}
   We propose a novel method that tackles the problem of unsupervised domain adaptation for semantic segmentation by maximizing the cosine similarity between the source and the target domain at the feature level. A segmentation network mainly consists of two parts, a feature extractor and a classification head. We expect that if we can make the two domains have small domain gap at the feature level, they would also have small domain discrepancy at the classification head. Our method computes a cosine similarity matrix between the source feature map and the target feature map, then we maximize the elements exceeding a threshold to guide the target features to have high similarity with the most similar source feature. Moreover, we use a class-wise source feature dictionary which stores the latest features of the source domain to prevent the unmatching problem when computing the cosine similarity matrix and be able to compare a target feature with various source features from various images. Through extensive experiments, we verify that our method gains performance on two unsupervised domain adaptation tasks (GTA5\textrightarrow Cityscaspes and SYNTHIA\textrightarrow Cityscapes).
   \vspace{-3mm}
\end{abstract}

\section{Introduction}
Semantic segmentation~\cite{long2015fully} is a pixel-wise classification task which segments an image based on semantic understanding. Recently, its progress has been significantly driven by deep convolutional neural networks. However, training a segmentation network requires dense pixel-level annotations which are laborious, time-consuming and expensive. Unsupervised Domain Adaptation (UDA) for semantic segmentation is one possible way to solve this problem. 
It adapts a model trained on a dataset with labels (source domain) to another dataset without labels (target domain). The source and the target domain datasets share common classes and environments, thus it is possible to adapt between the domains. Typically, as source domain datasets, synthetically generated computer graphic images such as GTA5 \cite{Richter_2016_ECCV} and SYNTHIA \cite{RosCVPR16} datasets which are relatively easy to get the annotations are used. On the other hand, a target domain dataset consists of real images, for example, Citycapses \cite{Cordts2016Cityscapes} dataset for which the annotations are hard to obtain. \par
\begin{figure}[t]
    \centering
    \includegraphics[width = 0.9\linewidth]{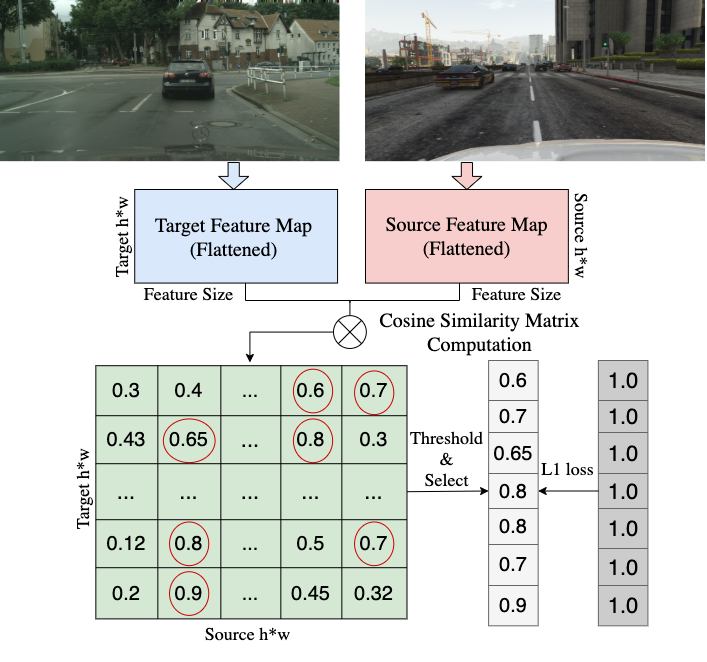}
    \caption{Overview of our cosine similarity loss at the feature level. By computing the cosine similarity matrix, we can compare which target feature is similar to which source feature and selectively maximize the cosine similarity between them spatial-wise. Here, the threshold is set to 0.5. Broad arrows on top indicate that the images are forwarded through the feature extractor of the segmentation network.}
    \label{fig:concept}
    \vspace{-3mm}
\end{figure}
Most of current UDA methods employ adversarial adaptation to overcome the domain discrepancy. It minimizes the discrepancy by fooling a discriminator network that is trained to distinguish the originating domain of an image correctly. However, it has a critical drawback that it only sees the output globally and checks whether it is from the source or the target domain, making it into a binary classification problem. This is not suitable to adapt target features to specific source features that are most similar. Wang et al. \cite{wang2020differential} pointed out these problems and showed that using adversarial loss can severely impair the performance in long-term training. Instead, they proposed a method of aligning the source and the target feature distributions by minimizing the L1 loss between the average feature representations of the two domains while treating the `stuff' and the `instance' classes differently. It is claimed that they could overcome the instability of adversarial adaptation and shift the target features towards the most similar source features. Nonetheless, this method has a couple of downsides that it computes an average of features for each class which loses spatial information and it still uses the adversarial adaptation. 

In this paper, we further investigate and develop in this direction of aligning the feature distributions by maximizing the cosine similarity between the two domains at the feature level.
Moreover, we show that adversarial adaptation is unnecessary and can be replaced by our newly-proposed method. Our intuition is that if the features of the two domains have high cosine similarity, their predictions would also be very similar. 
The contributions of our work are as follows. 
First, instead of taking an average of features for each class as in \cite{wang2020differential}, we compute a cosine similarity matrix to measure how a target feature is similar to each source feature spatial-wise. As can be seen in Fig.~\ref{fig:concept}, a cosine similarity matrix is computed between the flattened feature maps of the source and the target domain with respect to the feature dimension, producing a 2D matrix whose dimension is target's spatial size (height $\times$ width) by source's spatial size. Each row represents how a target feature is similar to each source feature along the spatial dimension. 
In practice, a feature map is split into classes so the cosine similarity matrix is computed for each class. From the matrix, we selectively maximize elements that have higher cosine similarity than a pre-defined threshold so that they become closer to 1. We believe that if a target feature is similar to a source feature, with their cosine similarity being higher than a certain threshold, those two features belong to the same semantic information (the same class). We call this as `cosine similarity loss' and it is inspired by contrastive learning~\cite{hadsell2006dimensionality, he2020momentum,chen2020simple,chen2020improved,grill2020bootstrap,chen2020big,tian2019contrastive,caron2020unsupervised} which pulls similar features closer and pushes dissimilar features apart. Our method nudges a target feature to the most similar source features of the same class. It suggests that even if the target and the source features belong to the same class, the target features have to be selectively closer to the source features that actually have high semantic similarity. 

Second, we use a dictionary that maintains the latest source features. Our method splits a source feature map by classes and stores them to a dictionary. The keys of the dictionary are the class identities and its values are the source features belonging to each class. The source features are stored as a queue, thus only the newest source features are kept for each class. 
The target feature map is also split by classes using either the pseudo-label or the prediction output. Then we compute the cosine similarity matrix between the split target features and the source features stored in the dictionary for each class. This approach enables to compare the target features with more variety of source features from various images. It also solves the unmatching problem which occurs when a certain class only appears in the current target image and not in the current source image thus the target features of the class do not have any source features to be maximized with.

For the last, we do not utilize the adversarial adaptation loss which is known to be complex and difficult to train and not appropriate to adapt target features to the most similar source features. We found that it is unnecessary and rather contradicts with our cosine similarity loss because it disrupts the training when used together. 
We empirically show that it can be replaced by our method. 
Therefore, we train only with the segmentation loss and our cosine similarity loss. We evaluate our method on two UDA of semantic segmentation benchmarks, `GTA5 $\rightarrow$ Cityscapes' and `SYNTHIA $\rightarrow$ Cityscapses' and show that our method has a valid performance gain.
\begin{figure*}[t]
    \centering
    \includegraphics[width = 1.0\linewidth]{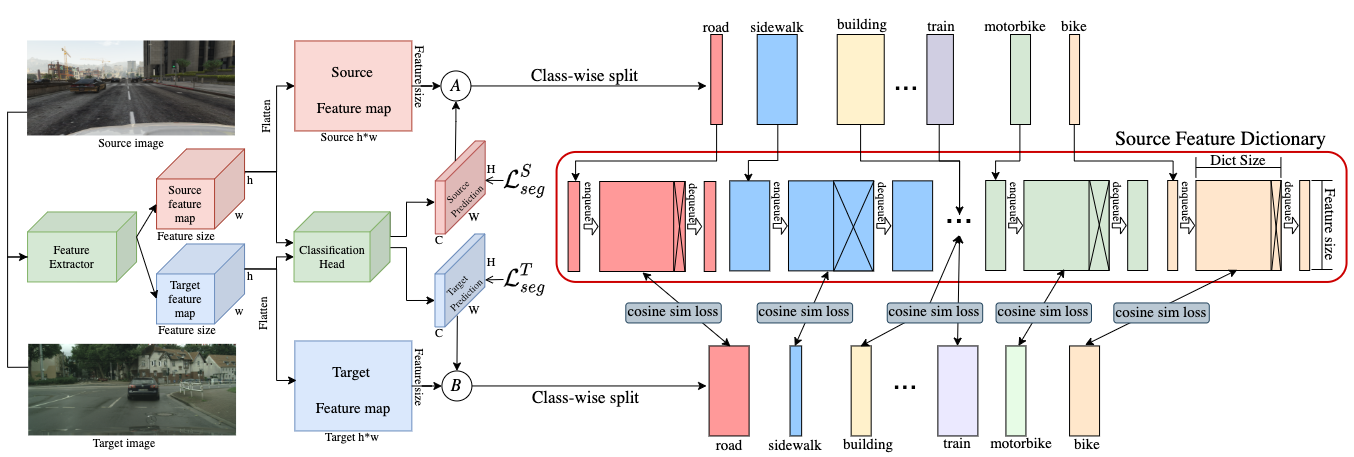}
    \caption{Overall schematic of our method. The feature extractor produces feature maps of both domains. Source and target feature maps are flattened and split by classes via process $\mathcal{A}$ and $\mathcal{B}$ respectively. Process $\mathcal{A}$ filters features that are correctly classified and splits them by classes. Process $\mathcal{B}$ splits the target feature map according to the pseudo-label or the prediction output. The source features are stored in the dictionary as a \textit{queue} to maintain the latest features for each class. Then the cosine similarity loss is computed class-wise between split target features and source features stored in the dictionary.
    }
    \label{fig:schematic}
    \vspace{-3mm}
\end{figure*}
\section{Related Work}
The goal of unsupervised domain adaptation (UDA) is to adapt and utilize the knowledge a model has learned from the source domain to perform well on the target domain without the supervision of target domain labels. This problem is challenging due to the discrepancy caused by the domain shift. UDA in classification is widely studied and has shown great progress \cite{zhang2019transfer}, on the other hand, UDA of semantic segmentation is more challenging sine it is a pixel-wise classification task.
Many works have been proposed in a variety of directions but we categorize them into three methodologies, image translation, adversarial adaptation and self-supervised learning via pseudo-labels. 

Image translation method closes the domain gap at the image level by style-transferring source domain images into target domain to maximize the visual similarity \cite{hoffman2018cycada, long2015fully, li2019bidirectional}. 
It tries to apply the style factors of the target domain to the source domain. Some works employ an image translation algorithm such as CycleGAN \cite{CycleGAN2017}. \cite{wu2018dcan} uses an image generator network to produce new synthesized source image via channel-wise feature alignment. \cite{musto2020semantically} proposes an adaptive image translation method which uses semantic output of the network. \cite{yang2020fda} shows a way to translate the source images without such complex translation network using Fourier transform and its inverse. \cite{kim2020learning} diversifies the texture of the source domain by style-transfer and train the segmentation network to learn texture-invariant representation. 

Adversarial adaptation tries to match the distribution of the source and the target domains at the feature and the prediction output level. It uses the adversarial training proposed in GAN \cite{goodfellow2014generative}. A  discriminator is employed to correctly distinguish from which domain the feature/output is generated from, while the segmentation network is trained to fool the discriminator. \cite{tsai2018learning} adopts multi-level adversarial training in the output space. \cite{vu2019advent} further improves this method and shows that using an entropy map of the prediction output produces better results. Adversarial adaptation is a very common method that is widely used in domain adaptation \cite{hoffman2016fcns,tzeng2017adversarial,shu2018dirt}. 

Self-supervised learning (SSL), or ``self-training" is a method of re-training a network with pseudo-labels of the target domain generated by a trained network that is adopted from the source domain. With the assistance of pseudo-labels, a network can be trained explicitly for the target domain as well as the source. \cite{zou2018domain} and \cite{li2019bidirectional} both utilize the pseudo-labels and propose a class-balanced and joint self-training respectively. \cite{pan2020unsupervised} tries to minimize the intra-domain gap via self-supervised adaptation by separating the target domain into an easy and a hard split. Recently, \cite{yang2020label} proposes a method that reconstructs the input-image from the network output to regularize the training of target domain along with self-supervised learning.

Our work adopts image translation and self-supervised learning while utilizing a novel cosine similarity loss at the feature level. We do not employ the adversarial adaptation since we find it to be ineffective, which will be empirically justified in Sec.\ref{wo-psuedo} and Sec.\ref{ablation_study}.

\section{Method}
In this section, we briefly talk about the loss function used for the semantic segmentation task and deeply investigate our proposed method. The schematic of our method is illustrated in Fig.~\ref{fig:schematic}. Our method utilizes a dictionary that stores the latest source features in a \textit{queue}. As described in Fig.~\ref{fig:concept}, we compute the class-wise cosine similarity between the target features and the source features stored in the dictionary. More detailed formulation of the proposed cosine similarity loss will be explained in Sec.~\ref{cos_sim_loss}. The cosine similarity loss is minimized along with the source and the target segmentation losses using the ground truth labels and the pseudo-labels respectively. Our cosine similarity loss pulls target features closer to the source features so that both domains are aligned in the feature space.

\subsection{Semantic Segmentation}
We follow the basic framework of unsupervised domain adaptation of semantic segmentation where there exist a source domain dataset with labels $\{x^s_i, y^s_i\}_{i=1}^{N_s}$ and a target domain dataset with only images $\{x^t_j\}_{j=1}^{N_t}$. Here, we assume that $y^s_i \in \mathbb{R}^{H\times W}$ with its elements being $y^{s(h,w)}_i \in [C]$\footnote{$[C]$ is the set of natural numbers up to $C$. For simplicity, the notations $y_i^s$ (label) and $\hat{y}_j^t$ (pseudo-label) are used to denote either an element of $[C]$ or a $C$-dimensional one-hot vector interchangeably in this paper.}. We train a segmentation network $\mathcal{G}$ that generates a prediction output $\mathcal{G}(x) = P \in \mathbb{R}^{H \times W \times C}$. We use the cross-entropy loss for the segmentation loss as follows:
\vspace{-3mm}
\begin{gather}
    \mathcal{L}_{seg}^S(x^s_i) = -\sum_{h,w}^{H,W} \sum_{c=1}^C {y}^{s(h,w,c)}_i\log (P_i^{s(h,w,c)}) \\
    \mathcal{L}_{seg}^T(x^t_j) = -\sum_{h,w}^{H,W} \sum_{c=1}^C \hat{{y}}^{t(h,w,c)}_j\log (P_j^{t(h,w,c)}) \\
    \mathcal{L}_{seg}(x^s_i, x^t_j) = \mathcal{L}_{seg}^S(x^s_i) + \mathcal{L}_{seg}^T(x^t_j)
    \label{Lseg}
    \vspace{-3mm}
\end{gather}
Here $H, W$ are height and width of the prediction output and $C$ denotes the number of classes. The source segmentation loss is defined using the ground truth labels provided by the source dataset, on the other hand, for the target segmentation loss, we adopt self-supervised learning scheme and use pseudo-labels denoted as $\{\hat{y}^t_j\}_{j=1}^{N_t}$ which are generated from a separate trained model. Following the process of \cite{li2019bidirectional}, only pixels with higher confidence than a threshold are filtered:
\vspace{-1mm}
\begin{equation}
    \begin{split}
    &(C^{t}_{max}, P^{t}_{max}) = (\argmax_{c\in [C]} P^{t(c)}, \max_{c\in [C]} P^{t(c)}) \\ 
    &\hat{y}^{t} = \mathbbm{1}_{[ P^{t}_{max} > \tau^{C^{t}_{max}}]} \odot C^{t}_{max} \in \mathbb{R}^{H \times W}
    \end{split}
    \label{eq:labelpseudo}
    \vspace{-3mm}
\end{equation}
where $\mathbbm{1}$ is an element-wise indicator function that returns 1 if the condition is met and 0 if not. $\odot$ denotes an element-wise multiplication. If $\hat{y}^{t(h,w)} = 0$, it indicates that the pixel $(h,w)$ is ignored. Detailed explanation for choosing the class-specific thresholds $\{\tau^c\}_{c \in [C]}$ is described in the supplementary.
The overall segmentation objective is to minimize $\mathcal{L}_{seg}$ to perform pixel-level classification for both domains.

\subsection{Cosine Similarity loss}
\label{cos_sim_loss}
The core idea of our method is to measure how similar the target features are to the source features spatial-wise and selectively maximize the similarity for certain target features that are highly similar to specific source features. First, we discuss how to split the source feature map by classes and store them to the source feature dictionary. A segmentation network, such as \cite{chen2017deeplab}, mainly consists of two parts, a feature extractor $\mathcal{F}$ and a classification head $\mathcal{H}$, hence $\mathcal{G} = \mathcal{H} \circ \mathcal{F}$. We feed a source image $x^s \in \mathbb{R}^{H \times W \times 3}$ into $\mathcal{F}$ and generate a feature map $f^s = \mathcal{F}(x^s) \in \mathbb{R}^{h \times w \times k}$ where $h$, $w$ and $k$ represent the height, width and the feature size (number of channels) of $f^s$. $\mathcal{H}$ takes $f^s$ and generates a prediction output, $p^s = \mathcal{H}(f^s)  \in \mathbb{R}^{h \times w \times C}$ followed by a bilinear interpolation $P^s = I_{biliner}(p^s) \in \mathbb{R}^{H \times W \times C}$. This can be put in one line as follows:
\vspace{-2mm}
\begin{equation}
    P^s = \mathcal{G}(x^s) = I_{biliner}(\mathcal{H}(\mathcal{F}(x^s))).
    \vspace{-2mm}
\end{equation}
We want to select correctly classified features from $f^s$ using $p^s$ and the ground truth label $y^s \in \mathbb{R}^{H \times W}$. We resize the ground truth label $y^s$ into the spatial size of $p^s$ via nearest interpolation, 
$\tilde{y}^s = I_{nearest}(y^s) \in \mathbb{R}^{h \times w}$.
\vspace{-2mm}
\begin{equation} 
\begin{split}
    c^s_{max} &= \argmax_{c\in [C]}p^{s(c)} \in \mathbb{R}^{h\times w}\\
    \hat{c}^s_{max} &= \mathbbm{1}_{[c^s_{max} = \tilde{y}^s]} \odot c^s_{max} \in \mathbb{R}^{h \times w}\\
    S^c  &= \mathbbm{1}_{[\hat{c}_{max}^s = c]} \otimes f^{s} \in \mathbb{R}^{h \times w \times k}.
    \vspace{-2mm}
    \label{eq:source-split}
\end{split}
\end{equation}
Here, $\otimes$ denotes the element-wise product of $\mathbbm{1}$ and each slice of $f^s$.
$\hat{c}^s_{max} $ contains information about the correctly classified output class according to $p^s$ and has the ignore symbol (0) where it is incorrectly classified. $S^c$ refers to the feature tensor $f^s$ that are correctly classified as class $c$ according to $\hat{c}_{max}^s$. Therefore, one $f^s$ can be split into maximum $C$ number of $S^c$. We flatten each $S^c$ along the spatial dimension, meaning that it has the shape of [$k \times {hw}^s_c$], where ${hw}^s_c$ refers to the number of pixels in $f^s$ that are correctly classified as $c$. Each $S^c$ that corresponds to each input $\{x_i^s\}_{i=1}^{N_s}$ is enqueued into the dictionary $D$ according to its class. $D$ has class identities as the keys and the values of each key are source features belonging to each class. $D^c$ refers to the values of $D$ accessed with key $c$ and it has the maximum size of \textit{dict-size} which is a hyper-parameter. $D$ is updated with new source features and the old features stored in $D$ are dequeued at every iteration. 

The reason we use the dictionary is to solve the case when a class appears only in the target image and not in the source image at the current iteration, we call this as the `unmatching problem'. In this case, the target features belonging to that class can not be matched with proper source features since the current source image does not contain the class. Also, using the dictionary allows the target features to be matched with more variety of source features from various images. $D^c$ has the shape of [$k \times \textit{dict-size}$] when it is fully queued. We call this process as $\mathcal{A}$.

Like the source features, we need to split a target feature map class-wise. Since target domain does not have ground truth labels, we consider two cases for splitting: when pseudo-labels are provided and when they are not. In the first case, we utilize the pseudo-labels. As in (\ref{eq:labelpseudo}), a pseudo-label $\hat{y}^t \in \mathbb{R}^{H \times W}$ has ignore symbols where the confidence of the trained model are lower than the threshold. Therefore, we augment $\hat{y}^t$ by replacing the ignore symbols with the prediction output of the current training network, $p^t \in \mathbb{R}^{h \times w \times C}$. We argmax $p^t$ along the class dimension and obtain $c_{max}^t \in \mathbb{R}^{h \times w}$. $\hat{y}^t$ is resized to the spatial size of $c_{max}^t$ as $\tilde{y}^t$ analogous to $\tilde{y}^s$.
\vspace{-2mm}
\begin{equation}
\begin{split}
    c_{max}^t = \argmax_{c\in [C]}p^{t(c)} \in \mathbb{R}^{h\times w}\\
    \dot{y}^t = augment(\tilde{y}^t, c_{max}^t) \in \mathbb{R}^{h \times w}\\
    T^c = \mathbbm{1}_{[\dot{y}^{t} = c]} \otimes f^{t} \in \mathbb{R}^{h \times w \times k}.
    \label{split-pseudo}
    \vspace{-2mm}
\end{split}
\end{equation}
We augment $\tilde{y}^t$ with $c_{max}^t$, generating $\dot{y}^t_j$ which has the values of $c_{max}^t$ where $\tilde{y}^t$ has ignore symbols. We split a target feature map $f^t$ according to the augmented pseudo-label $\dot{y}^t$.

In the second case, we split a target feature map only according to $c_{max}^t$.
\vspace{-2mm}
\begin{equation}
\begin{split}
    T^c = \mathbbm{1}_{[c_{max}^t = c]} \otimes f^{t} \in \mathbb{R}^{h \times w \times k}.
    \label{split-no-pseudo}
    \vspace{-2mm}
\end{split}
\end{equation}
Therefore, $f^t$ can be split into maximum $C$ number of $T^c$ same as $f^s$. $T^c$ is flattened along the spatial dimension hence its shape is [$k \times {hw}^t_c$] analogous to $S^c$.
We name this process as $\mathcal{B}$. 
The process $\mathcal{A}$ and $\mathcal{B}$ are illustrated in the supplementary.

\begin{figure}[t]
    \centering
    \includegraphics[width = 1.0\linewidth]{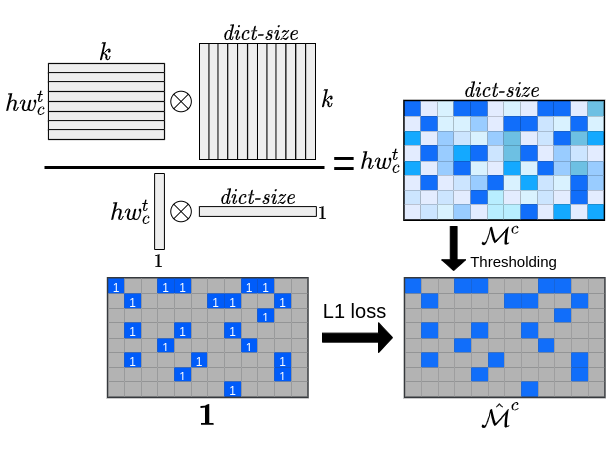}
    \vspace{-2mm}
    \caption{Dark grey colored parts refer to ignore symbols. The different blue colors for the elements of $\mathcal{M}^c$ refer to different scales of cosine similarity. Darker means higher cosine similarity.}
    \label{fig:cosine-sim-matrix}
    \vspace{-2mm}
\end{figure}
Next step is to compute a cosine similarity matrix between split target features and source features stored in the dictionary for each class (Also, see Fig. \ref{fig:cosine-sim-matrix}). 
\vspace{-2mm}
\begin{equation}
    \mathcal{M}^c = \frac{{T^c}^\mathsf{T} \boldsymbol{\cdot} D^c}{\norm{T^c}^\mathsf{T}_2 \boldsymbol{\cdot} \norm{D^c}_2}
\end{equation}
$\norm{T^c}_2$ and $\norm{D^c}_2$ are L2-norms of ${T^c}$ and $D^c$ with respect to the feature dimension, thus their shapes are [$1 \times {hw}^t_c$] and [${1 \times \textit{dict-size}}$], assuming that $D^c$ is fully queued. 
$\mathsf{T}$ and $\boldsymbol{\cdot}$ are transpose and matrix multiplication. Therefore, $\mathcal{M}^c$ has the shape of [${hw}^t_c \times \textit{dict-size}$] which shows the cosine similarity between current target features and the source features stored in the dictionary for class $c$. Every element of $\mathcal{M}^c$ is a cosine similarity thus each row represents how one target feature vector is similar to every source feature vector stored in the $D^c$. Since cosine similarity is normalized between 1 and $-1$, the closer the element is to 1, the more similar the two feature vectors of target and source are. We then select elements of $\mathcal{M}^c$ that exceeds the threshold $\mathcal{T}_{cos}$ and maximize those elements to 1. We detach $D^c$ when computing $\mathcal{M}^c$, so the loss does not back-propagate through $D^c$ but only through ${T^c}$.
\vspace{-5mm}
\begin{equation}
\begin{split}
    \hat{\mathcal{M}}^c = \mathbbm{1}_{[\mathcal{M}^c>\mathcal{T}_{cos}]} \odot \mathcal{M}^c \\
    \mathcal{L}_{cos}(x^t) =\frac{1}{C} \sum_{c=1}^C\norm{\hat{\mathcal{M}}^c - \textbf{1}}^1_1
    \vspace{-5mm}
\end{split}
\end{equation}
$\hat{\mathcal{M}}^c$ has selected elements of $\mathcal{M}^c$ that have higher cosine similarity than $\mathcal{T}_{cos}$. $\textbf{1}$ has the same shape as $\hat{\mathcal{M}}^c$ but filled with 1 where the elements of $\hat{\mathcal{M}}^c$ are not ignore symbols. We maximize the selected elements to 1 by minimizing L1 loss between $\hat{\mathcal{M}}^c$ and $\textbf{1}$. Fig.~\ref{fig:cosine-sim-matrix} gives an illustration of computing the cosine similarity loss for class $c$. 
Because our method compares a target feature with every source feature in $D^c$, it can selectively maximize the similarity of target features to the most similar source features, making the target feature distribution and the source feature distribution aligned in the feature space.

The total loss function is to minimize the segmentation loss along with our cosine similarity loss with balance parameter $\lambda_{cos}$. (\ref{step1}) is used when the pseudo-labels for the target domain are \textbf{not} provided thus the target feature map is split using (\ref{split-no-pseudo}). (\ref{step2}) is when the pseudo-labels are available thus the target feature map is split by (\ref{split-pseudo}). Note that we do not employ any adversarial adaptation loss.
\begin{equation}
    \mathcal{L}_{total}(x^s_i, x^t_j) = \mathcal{L}_{seg}^S(x^s_i) + \lambda_{cos}\mathcal{L}_{cos}(x^t_j) \\
    \label{step1}
\end{equation}
\begin{equation}
    \mathcal{L}_{total}(x^s_i, x^t_j) = \mathcal{L}_{seg}(x^s_i, x^t_j) + \lambda_{cos}\mathcal{L}_{cos}(x^t_j).
    \label{step2}
\end{equation}

\section{Experiments}
\subsection{Datasets and Training Details}
\textbf{Datasets. } We conduct experiments on two UDA benchmarks, GTA5\textrightarrow Cityscapes and SYNTHIA\textrightarrow Cityscapes. The GTA5 \cite{Richter_2016_ECCV} dataset consists of 24,966 images captured from a video game with pixel-level annotations. Originally it has annotations for 33 classes, but only 19 classes that are in common with Cityscapes are used to fairly compare with other methods. Images are resized to 1280 $\times$ 720 during training. The SYNTHIA \cite{RosCVPR16} dataset also consists of 9,400 synthetic images with a resolution of 1280 $\times$ 760. Similar to GTA5, 16 common classes with Cityscapes are used for training, but for evaluation, the 16 classes and a subset with 13 classes are used following the standard protocol. The Cityscapes \cite{Cordts2016Cityscapes} dataset is a semantic segmentation dataset collected from real world during driving scenarios. We use 2,975 images of the train set to train the model and 500 images of the validation set to test our model, following previous works. Images are resized to 1024 $\times$ 512 during training.

\textbf{Network Architecture and Training Details.} We use two different network architectures, DeepLabV2 \cite{chen2017deeplab} with ResNet101 \cite{he2016deep} backbone and FCN-8s \cite{long2015fully} with VGG16 backbone \cite{simonyan2014very}. Both networks are initialized with an ImageNet pre-trained networks. We do not employ any discriminator network so the segmentation network is the only neural network in usage. Pytorch deep learning framework is used on single GPU. Batch size is set to 1 due to limited memory, same as other methods~\cite{vu2019advent,li2019bidirectional,yang2020fda,wang2020differential}. For DeepLabV2 with ResNet101 backbone, we use SGD as the optimizer with initial learning rate of $2.5 \times 10^{-4}$ and weight-decay of 0.0005. Learning rate is scheduled using `poly' learning rate policy with a power of 0.9. FCN-8s with VGG16 backbone is optimized by ADAM optimizer with an initial learning rate of $1 \times 10^{-5}$ and the momentums of 0.9 and 0.99. The learning rate is decayed by 'step' learning rate policy with a step size of 50,000 and a decay rate of 0.1. We adopt the transferred source images of \cite{li2019bidirectional} which are transferred into the style of Cityscapes by CycleGAN \cite{CycleGAN2017}.
Hyper-parameters used in our method such as $\mathcal{T}_{cos}$, $\textit{dict-size}$ and $\lambda_{cos}$ will be discussed in Sec.~\ref{hyper-params}.

\begin{table}[t]
		\begin{center}
		    \resizebox{0.9\linewidth}{!}{
			\begin{tabular}{l|l|c|c}
				\toprule
				Architecture & Method & GTA5\textrightarrow CS & SYNTHIA\textrightarrow CS \\ 
				\midrule
				{\multirow{2}{*}{DeeplabV2}}
				& Adversarial & 45.1\% & 48.5\% \\
				 & Ours-(\ref{step1}) & 46.6\% & 48.6\% \\ 
				\midrule
				{\multirow{2}{*}{FCN-8s}}
				& Adversarial & 40.7\% & 36.2\% \\
				 & Ours-(\ref{step1}) & 41.8\% & 36.5\% \\
				\bottomrule
			\end{tabular}}
		\end{center}
		\vspace{-2mm}
		\caption{The results of training without the pseudo-labels for both tasks using two different network architectures.}
		\label{table:wo-pseudo}
		\vspace{-3mm}
\end{table}

\subsection{Training without pseudo-labels}
\label{wo-psuedo}
Tab.~\ref{table:wo-pseudo} shows our results of training without pseudo-labels. When the pseudo-labels are not provided, the adversarial adaptation loss is usually utilized to match the distribution of the target prediction to that of the source prediction. However, as mentioned earlier, adversarial adaptation is difficult to train and requires an additional discriminator network. Moreover, It is known to cause instability in the long-term training since it only sees the global prediction output and not the details.
`Adversarial' in the table refers to adversarial adaptation method \cite{tsai2018learning} which tries to align the distributions of the two domains at the prediction output level using the adversarial training. Detailed formulation of it is in the supplementary. `Ours-(\ref{step1})' refers to models trained with our cosine similarity loss using (\ref{step1}) which splits the target feature map solely based on the target prediction output. By comparing the results of `Ours-(\ref{step1})' with `Adversarial', we want to show that, when pseudo-labels are unavailable, our method can not only replace the adversarial adaptation but also lead to better performance results. Overall, Tab.~\ref{table:wo-pseudo} indicates that with our cosine similarity loss, adversarial adaptation loss is unnecessary and can be replaced. 

\begin{table*}[t]
		\setlength\tabcolsep{0.15em}
		\begin{center}
		    \resizebox{0.88\textwidth}{!}{
			\begin{tabular}{ @{} l|c|*{19}{c}|*{1}{c} @{} }
				\toprule
				& \rotatebox[origin=c]{50}{Arch.} & \rotatebox[origin=c]{50}{road} & \rotatebox[origin=c]{50}{sidewalk} & \rotatebox[origin=c]{50}{building} & \rotatebox[origin=c]{50}{wall} & \rotatebox[origin=c]{50}{fence} & \rotatebox[origin=c]{50}{pole} & \rotatebox[origin=c]{50}{  traffic light  } & \rotatebox[origin=c]{50}{traffic sign} & \rotatebox[origin=c]{50}{vegetation} & \rotatebox[origin=c]{50}{terrain} & \rotatebox[origin=c]{50}{sky} & \rotatebox[origin=c]{50}{person} & \rotatebox[origin=c]{50}{rider} & \rotatebox[origin=c]{50}{car} & \rotatebox[origin=c]{50}{truck} & \rotatebox[origin=c]{50}{bus} & \rotatebox[origin=c]{50}{train} & \rotatebox[origin=c]{50}{motorbike} & \rotatebox[origin=c]{50}{bicycle} & \rotatebox[origin=c]{50}{\bf mIoU } \\ 
				\midrule
				
				AdaStruct\cite{tsai2018learning} & R & 86.5 & 36.0 & 79.9 & 23.4 & 23.3 & 23.9 & 35.2 & 14.8 & 83.4 & 33.3 & 75.6 & 58.5 & 27.6 & 73.7 & 32.5 & 35.4 & 3.9 & 30.1 & 28.1 & 42.4 \\
				
				SIBAN \cite{luo2019significance} & R & 88.5 & 35.4 & 79.5 & 26.3 & 24.3 & 28.5 & 32.5 & 18.3 & 81.2 & 40.0 & 76.5 & 58.1 & 25.8 & 82.6 & 30.3 & 34.4 & 3.4 & 21.6 & 21.5 & 42.6 \\ 
				
				CyCADA\cite{hoffman2018cycada} & R & 86.7 & 35.6 & 80.1 & 19.8 & 17.5 & 38.0 & 39.9 & 41.5 & 82.7 & 27.9 & 73.6 & 64.9 & 19 & 65.0 & 12.0 & 28.6 & 4.5 & 31.1 & 42.0 & 42.7 \\
				
				CLAN \cite{luo2019taking} & R & 87.0 & 27.1 & 79.6 & 27.3 & 23.3 & 28.3 & 35.5 & 24.2 & 83.6 & 27.4 & 74.2 & 58.6 & 28.0 & 76.2 & 33.1 & 36.7 & 6.7 & 31.9 & 31.4 & 43.2 \\
				
				DISE \cite{chang2019all} & R & 91.5 & 47.5 & 82.5 & 31.3 & 25.6 & 33.0 & 33.7 & 25.8 & 82.7 & 28.8 & 82.7 & 62.4 & 30.8 & 85.2 & 27.7 & 34.5 & 6.4 & 25.2 & 24.4 & 45.4 \\
				
				AdvEnt\cite{vu2019advent} & R & 89.4 & 33.1 & 81.0 & 26.6 & 26.8 & 27.2 & 33.5 & 24.7 & 83.9 & 36.7 & 78.8 & 58.7 & 30.5 & 84.8 & 38.5 & 44.5 & 1.7 & 31.6 & 32.4 & 45.5 \\
				
			    IntraDA \cite{pan2020unsupervised} & R & 90.6 & 37.1 & 82.6 & 30.1 & 19.1 & 29.5 & 32.4 & 20.6 & 85.7 & 40.5 & 79.7 & 58.7 & 31.1 & 86.3 & 31.5 & 48.3 & 0.0 & 30.2 & 35.8 & 46.3  \\
				
				BDL \cite{li2019bidirectional} & R & 91.0 & 44.7 & 84.2 & 34.6 & 27.6 & 30.2 & 36.0 & 36.0 & 85.0 & 43.6 & 83.0 & 58.6 & 31.6 & 83.3 & 35.3 & 49.7 & 3.3 & 28.8 & 35.6 & 48.5 \\
				
				CrCDA \cite{huang2020contextual}  & R & 92.4 & 55.3 & 82.3 & 31.2 & 29.1 & 32.5 & 33.2 & 35.6 & 83.5 & 34.8 & 84.2 & 58.9 & 32.2 & 84.7 & 40.6 & 46.1 & 2.1 & 31.1 & 32.7 & 48.6 \\
				
				SIM \cite{wang2020differential} & R & 90.6 & 44.7 & 84.8 & 34.3 & 28.7 & 31.6 & 35.0 & 37.6 & 84.7 & 43.3 & 85.3 & 57.0 & 31.5 & 83.8 & 42.6 & 48.5 & 1.9 & 30.4 & 39.0 & 49.2 \\
				
				Label-driven\cite{yang2020label} & R &
				90.8 & 41.4 & 84.7 & 35.1 & 27.5 & 31.2 & 38.0 & 32.8&85.6&42.1&84.9&59.6&34.4&85.0&42.8&52.7&3.4&30.9&38.1&49.5 \\
				
				Kim et al. \cite{kim2020learning} & R & 92.9 & 55.0 & 85.3 & 34.2 & 31.1 & 34.9 & 40.7 & 34.0 & 85.2 & 40.1 & 87.1 & 61.0 & 31.1 & 82.5 & 32.3 & 42.9 & 0.3 & 36.4 & 46.1 & 50.2 \\
				
				FDA-MBT \cite{yang2020fda} & R & 92.5 & 53.3 & 82.4 & 26.5 & 27.6 & 36.4 & 40.6 & 38.9 & 82.3 & 39.8 & 78.0 & 62.6 & 34.4 & 84.9 & 34.1 & 53.1 & 16.9 & 27.7 & 46.4 & \bf 50.45 \\
				
				\midrule
				Ours & R & 92.6 & 54.0 & 85.4 & 35.0 & 26.0 & 32.4 & 41.2 & 29.7 & 85.1 & 40.9 & 85.4 & 62.6 & 34.7 & 85.7 & 35.6 & 50.8 & 2.4 & 31.0 & 34.0 & 49.7 \\
				
				\midrule
				SIBAN \cite{luo2019significance} & V & 83.4 & 13.0 & 77.8 & 20.4 & 17.5 & 24.6 & 22.8 & 9.6 & 81.3 & 29.6 & 77.3 & 42.7 & 10.9 & 76.0 & 22.8 & 17.9 & 5.7 & 14.2 & 2.0 & 34.2 \\
				
				AdaStruct \cite{tsai2018learning} & V & 87.3 & 29.8 & 78.6 & 21.1 & 18.2 & 22.5 & 21.5 & 11.0 & 79.7 & 29.6 & 71.3 & 46.8 & 6.5 & 80.1 & 23.0 & 26.9 & 0.0 & 10.6 & 0.3 & 35.0 \\
				
				CyCADA \cite{hoffman2018cycada} & V & 85.2 & 37.2 & 76.5 & 21.8 & 15.0 & 23.8 & 22.9 & 21.5 & 80.5 & 31.3 & 60.7 & 50.5 & 9.0 & 76.9 & 17.1 & 28.2 & 4.5 & 9.8 & 0.0 & 35.4 \\    
				
				AdvEnt\cite{vu2019advent} & V & 86.9 & 28.7 & 78.7 & 28.5 & 25.2 & 17.1 & 20.3 & 10.9 & 80.0 & 26.4 & 70.2 & 47.1 & 8.4 & 81.5 & 26.0 & 17.2 & 18.9 & 11.7 & 1.6 & 36.1 \\
				
				CLAN \cite{luo2019taking} & V & 88.0 & 30.6 & 79.2 & 23.4 & 20.5 & 26.1 & 23.0 & 14.8 & 81.6 & 34.5 & 72.0 & 45.8 & 7.9 & 80.5 & 26.6 & 29.9 & 0.0 & 10.7 & 0.0 & 36.6 \\
				
				CrDoCo \cite{chen2019crdoco} & V & 89.1 & 33.2 & 80.1 & 26.9 & 25.0 & 18.3 & 23.4 & 12.8 & 77.0 & 29.1 & 72.4 & 55.1 & 20.2 & 79.9 & 22.3 & 19.5 & 1.0 & 20.1 & 18.7 & 38.1 \\
				
				CrCDA \cite{huang2020contextual} & V & 86.8 & 37.5 & 80.4 & 30.7 & 18.1 & 26.8 & 25.3 & 15.1 & 81.5 & 30.9 & 72.1 & 52.8 & 19.0 & 82.1 & 25.4 & 29.2 & 10.1 & 15.8 & 3.7 & 39.1 \\
				
				BDL \cite{li2019bidirectional} & V & 89.2 & 40.9 & 81.2 & 29.1 & 19.2 & 14.2 & 29.0 & 19.6 & 83.7 & 35.9 & 80.7 & 54.7 & 23.3 & 82.7 & 25.8 & 28.0 & 2.3 & 25.7 & 19.9 & 41.3 \\  
				
				FDA-MBT \cite{yang2020fda} & V & 86.1 & 35.1 & 80.6 & 30.8 & 20.4 & 27.5 & 30.0 & 26.0 & 82.1 & 30.3 & 73.6 & 52.5 & 21.7 & 81.7 & 24.0 & 30.5 & 29.9 & 14.6 & 24.0 & 42.2 \\
				
				Kim et al. \cite{kim2020learning} & V & 92.5 & 54.5 & 83.9 & 34.5 & 25.5 & 31.0 & 30.4 & 18.0 & 84.1 & 39.6 &  83.9 & 53.6 & 19.3 & 81.7 & 21.1 & 13.6 & 17.7 & 12.3 & 6.5 & 42.3 \\
				
				SIM \cite{wang2020differential} & V & 88.1 & 35.8 & 83.1 & 25.8 & 23.9 & 29.2 & 28.8 & 28.6 & 83.0 & 36.7 & 82.3 & 53.7 & 22.8 & 82.3 & 26.4 & 38.6 & 0.0 & 19.6 & 17.1 & 42.4 \\
				
				Label-driven\cite{yang2020label} & V & 90.1&41.2&82.2&30.3&21.3&18.3&33.5&23.0&84.1&37.5&81.4&54.2&24.3&83.0&27.6&32.0&8.1&29.7&26.9&43.6 \\    
				
				\midrule
				Ours & V & 90.3 & 42.6 & 82.2 & 29.7 & 22.2 & 18.5 & 32.8 & 26.8 & 84.3 & 37.1 & 80.2 & 55.2 & 26.4 & 83.0 & 30.3 & 35.1 & 7.0 & 29.6 & 28.9 & \bf 44.3 \\    
				
				\bottomrule
			\end{tabular}}
		\end{center}
		\vspace{-2mm}
		\caption{Performance comparison of our method with other state-of-the-art methods on \textbf{GTA5\textrightarrow Cityscapes}. 'R' and 'V' refer to DeepLabV2-ResNet101 and FCN-8s-VGG16 respectively.}
		\label{table:gta2city}
		\vspace{-2mm}
\end{table*}
\begin{table*}[t]
		\setlength\tabcolsep{0.15em}
		\begin{center}
		    \resizebox{0.88\textwidth}{!}{
			\begin{tabular}{ @{} l|c|*{16}{c}|*{1}{c} @{} }
				\toprule
				& \rotatebox[origin=c]{50}{Arch.} & \rotatebox[origin=c]{50}{road} & \rotatebox[origin=c]{50}{sidewalk} & \rotatebox[origin=c]{50}{building} & \rotatebox[origin=c]{50}{wall} & \rotatebox[origin=c]{50}{fence} & \rotatebox[origin=c]{50}{pole} & \rotatebox[origin=c]{50}{traffic light} & \rotatebox[origin=c]{50}{traffic sign} & \rotatebox[origin=c]{50}{vegetation} & \rotatebox[origin=c]{50}{sky} & \rotatebox[origin=c]{50}{person} & \rotatebox[origin=c]{50}{rider} & \rotatebox[origin=c]{50}{car} & \rotatebox[origin=c]{50}{bus} & \rotatebox[origin=c]{50}{motorbike} & \rotatebox[origin=c]{50}{bicycle} & \rotatebox[origin=c]{50}{\bf mIoU} \\ 
				\midrule
				
				AdaStruct \cite{tsai2018learning} & R & 84.3 & 42.7 & 77.5 & \textemdash & \textemdash & \textemdash & 4.7 & 7.0 & 77.9 & 82.5 & 54.3 & 21.0 & 72.3 & 32.2 & 18.9 & 32.3 & 46.7 \\
				
				CLAN \cite{luo2019taking} & R & 81.3 & 37.0 & 80.1 & \textemdash & \textemdash & \textemdash & 16.1 & 13.7 & 78.2 & 81.5 & 53.4 & 21.2 & 73.0 & 32.9 & 22.6 & 30.7 & 47.8 \\
				
				AdvEnt\cite{vu2019advent} & R & 85.6 & 42.2 & 79.7 & \textemdash & \textemdash & \textemdash & 5.4 & 8.1 & 80.4 & 84.1 & 57.9 & 23.8 & 73.3 & 36.4 & 14.2 & 33.0 & 48.0 \\
				
				DISE \cite{chang2019all} & R & 91.7 & 53.5 & 77.1 & \textemdash & \textemdash & \textemdash & 6.2 & 7.6 & 78.4 & 81.2 & 55.8 & 19.2 & 82.3 & 30.3 & 17.1 & 34.3 & 48.8\\
				
				IntraDA \cite{pan2020unsupervised} & R & 84.3 & 37.7 & 79.5 & \textemdash & \textemdash & \textemdash & 9.2 & 8.4 & 80.0 & 84.1 & 57.2 & 23.0 & 78.0 & 38.1 & 20.3 & 36.5 & 48.9 \\
				
				Kim et al. \cite{kim2020learning} & R & 92.6 & 53.2 & 79.2 & \textemdash & \textemdash & \textemdash & 1.6 & 7.5 & 78.6 & 84.4 & 52.6 & 20.0 & 82.1 & 34.8 & 14.6 & 39.4 & 49.3 \\
				
				DADA \cite{vu2019dada} & R & 89.2 & 44.8 & 81.4 & \textemdash & \textemdash & \textemdash & 8.6 & 11.1 & 81.8 & 84.0 & 54.7 & 19.3 & 79.7 & 40.7 & 14.0 & 38.8 & 49.8 \\
				
				CrCDA \cite{huang2020contextual} & R & 86.2 & 44.9 & 79.5 & \textemdash & \textemdash & \textemdash & 9.4 & 11.8 & 78.6 & 86.5 & 57.2 & 26.1 & 76.8 & 39.9 & 21.5 & 32.1 & 50.0 \\
				
				BDL \cite{li2019bidirectional} & R &
				86.0 & 46.7 & 80.3 & \textemdash & \textemdash & \textemdash & 14.1 & 11.6 & 79.2 & 81.3 & 54.1 & 27.9 & 73.7 & 42.2 & 25.7 & 45.3 & 51.4 \\
				
				SIM \cite{wang2020differential} & R &
				83.0 & 44.0 & 80.3 & \textemdash & \textemdash &  \textemdash & 17.1 & 15.8 & 80.5 & 81.8 & 59.9 & 33.1 & 70.2 & 37.3 & 28.5 & 45.8 & 52.1 \\
				
				FDA-MBT \cite{yang2020fda} & R & 79.3 & 35.0 & 73.2 & \textemdash & \textemdash &  \textemdash & 19.9 & 24.0 & 61.7 & 82.6 & 61.4 & 31.1 &  83.9 & 40.8 & 38.4 & 51.1 & 52.5 \\
				
				Label-driven\cite{yang2020label} & R & 85.1 & 44.5 & 81.0 & \textemdash & \textemdash & \textemdash & 16.4 & 15.2 & 80.1 & 84.8 & 59.4 & 31.9 & 73.2 & 41.0 & 32.6 & 44.7 & 53.1 \\
				
				\midrule
				Ours & R & 88.3 & 47.3 & 80.1 & \textemdash & \textemdash & \textemdash& 21.6 & 20.2 & 79.6 & 82.1 & 59.0 & 28.2 & 82.0 & 39.2 & 17.3 & 46.7 & \bf53.2 \\
				
				\midrule
		        AdvEnt\cite{vu2019advent} & V & 67.9 & 29.4 & 71.9 & 6.3 & 0.3 & 19.9 & 0.6 & 2.6 & 74.9 & 74.9 & 35.4 & 9.6 & 67.8 & 21.4 & 4.1 & 15.5 & 31.4 \\
		        
				CrCDA \cite{huang2020contextual} & V & 74.5 & 30.5 & 78.6 & 6.6 & 0.7 & 21.2 & 2.3 & 8.4 & 77.4 & 79.1 & 45.9 & 16.5 & 73.1 & 24.1 & 9.6 & 14.2 & 35.2 \\
				
				ROAD-Net \cite{chen2018road} & V & 77.7 & 30.0 & 77.5 & 9.6 & 0.3 & 25.8 & 10.3 & 15.6 & 77.6 & 79.8 & 44.5 & 16.6 & 67.8 & 14.5 & 7.0 & 23.8 & 36.2 \\
				
				SPIGAN \cite{lee2018spigan} & V & 71.1 & 29.8 & 71.4 & 3.7 & 0.3 & 33.2 & 6.4 & 15.6 & 81.2 & 78.9 & 52.7 & 13.1 & 75.9 & 25.5 & 10.0 & 20.5 & 36.8 \\
				
				GIO-Ada \cite{chen2019learning} & V & 78.3 & 29.2 & 76.9 & 11.4 & 0.3 & 26.5 & 10.8 & 17.2 & 81.7 & 81.9 & 45.8 & 15.4 & 68.0 & 15.9 & 7.5 & 30.4 & 37.3 \\
				
				CrDoCo \cite{chen2019crdoco} & V & 84.9 & 32.8 & 80.1 & 4.3 & 0.4 & 29.4 & 14.2 & 21.0 & 79.2 & 78.3 & 50.2 & 15.9 & 69.8 & 23.4 & 11.0 & 15.6 & 38.2 \\
				
				TGCF-DA \cite{Choi2019self} & V & 90.1 & 48.6 & 80.7 & 2.2 & 0.2 & 27.2 & 3.2 & 14.3 & 82.1 & 78.4 & 54.4 & 16.4 & 82.5 & 12.3 & 1.7 & 21.8 & 38.5 \\
				
				BDL \cite{li2019bidirectional} & V & 72.0 & 30.3 & 74.5 & 0.1 & 0.3 & 24.6 & 10.2 & 25.2 & 80.5 & 80.0 & 54.7 & 23.2 & 72.7 & 24.0 & 7.5 & 44.9 & 39.0 \\
				
				FDA-MBT \cite{yang2020fda} & V & 84.2 & 35.1 & 78.0 & 6.1 & 0.44 & 27.0 & 8.5 & 
				22.1 & 77.2 & 79.6 & 55.5 & 19.9 & 74.8 & 24.9 & 14.3 & 40.7 & 40.5 \\
				
				Label-driven\cite{yang2020label} & V & 73.7&29.6&77.6&1.0&0.4&26.0&14.7&26.6&80.6&81.8&57.2&24.5&76.1&27.6&13.6&46.6&
				41.1 \\
				
				\midrule
				Ours & V & 73.6 & 30.6 & 77.5 & 0.8 & 0.4 & 26.7 & 14.1 & 29.3 & 80.9 & 80.6 & 57.9 & 24.7 & 76.5 & 27.2 & 10.8 & 47.8 &\bf41.2 \\
				
				\bottomrule
			\end{tabular}}
		\end{center}
		\vspace{-2mm}
		\caption{Performance comparison of our method with other state-of-the-art methods on \textbf{SYNTHIA\textrightarrow Cityscapes}. 'R' and 'V' refer to DeepLabV2-ResNet101 and FCN-8s-VGG16 respectively. mIoU13 and mIoU16 are used for 'R' and 'V'.}
		\label{table:synthia2city}
		\vspace{-2mm}
\end{table*}

\subsection{Comparison with other SOTA methods}
In this section, we compare our results with other state-of-the-art methods on GTA5\textrightarrow Cityscapes and SYNTHIA\textrightarrow Cityscapes using two network architectures, DeepLabV2 based on ResNet101 and FCN-8s based on VGG16. Our results are models trained by (\ref{step2}). We use the pseudo-labels generated from models trained by \cite{vu2019advent} and \cite{li2019bidirectional} for DeepLabV2 and FCN-8s respectively. Considering that FCN-8s has a different architecture from DeepLabV2, our method is tweaked a little to adapt the difference. We basically use the cosine similarity loss for three different layers, fc7, pool4 and pool3, and more explanation is in the supplementary. We use mIoU of all 19 classes for GTA5\textrightarrow Cityscapes, but for SYNTHIA\textrightarrow Cityscapes, mIoU13 and mIoU16 are used for DeepLabV2 and FCN-8s respectively, following the standard evaluation protocol.

\textbf{GTA5\textrightarrow Cityscapes.} Tab.~\ref{table:gta2city} shows our comparison experiment on GTA5\textrightarrow Cityscapes. For DeepLabV2-ResNet101 model, our method outperforms most of current UDA methods except for \cite{kim2020learning} and \cite{yang2020fda}. However, these two approaches are very different from ours which are to style-transfer source images into many different textures and Cityscapes respectively. It is more about how to style-transfer the source images properly to adapt well rather than aligning the feature distributions of the source and the target domains. Furthermore, the mIoU of \cite{yang2020fda} is an ensemble average of three different models, not a single model, whose best single model mIoU (48.77\%) is below ours. Since these two works have different directions from ours, we expect better results can be produced when our method is combined with them. Our method achieves the best results for `person' and `rider' which are important classes to detect under autonomous driving scenario for safety. For FCN-8s-VGG16 model, our method outperforms other existing methods and achieves the new state-of-the-art.

\textbf{SYNTHIA\textrightarrow Cityscapes.} Tab.~\ref{table:synthia2city} compares our method with other methods on SYNTHIA\textrightarrow Cityscapes. SYNTHIA\textrightarrow Cityscapes is a more difficult task than GTA5\textrightarrow Cityscapes since the domain discrepancy is much larger. The images in SYNTHIA dataset have different perspectives from Citysacpses. There are more viewpoints from a higher position such as traffic surveillance cameras. Despite the difficulty of the dataset, as can be seen in the table, our method outperforms other existing methods and achieves the new state-of-the-art for both network architectures. 

\begin{table}[t]
    \centering
    \small
    \resizebox{1.0\linewidth}{!}{
    \begin{tabular}{l|l|c|c}
        \toprule
         Arch. & Method & GTA5\textrightarrow CS & SYNTHIA\textrightarrow CS  \\
         \midrule
         {\multirow{6}{*}{R}}
         & Ours & 49.7\% & 53.2\% \\
         & w/o Dictionary & 49.33\% & 52.64\% \\
         & w/o Class-wise Split & 48.93\% & 52.67\% \\
         & with Adversarial & 48.71\% & 52.75\% \\
         & only SSL & 48.65\% & 52.48\% \\
         & SSL with Adversarial & 48.41\% & 52.56\% \\
         \midrule
         {\multirow{2}{*}{V}}
         & Ours & 44.3\% & 41,2\% \\
         & only SSL & 43.52\% & 40.81\% \\
         \bottomrule
    \end{tabular}}
    \caption{The ablation study results. Note that 'R' and 'V' refer to DeepLabV2-ResNet101 and FCN-8s-VGG16 respectively.}
    \label{table:ablation}
    \vspace{-3mm}
\end{table}

\begin{figure}[t]
    \centering
    \includegraphics[width = 1.0\linewidth]{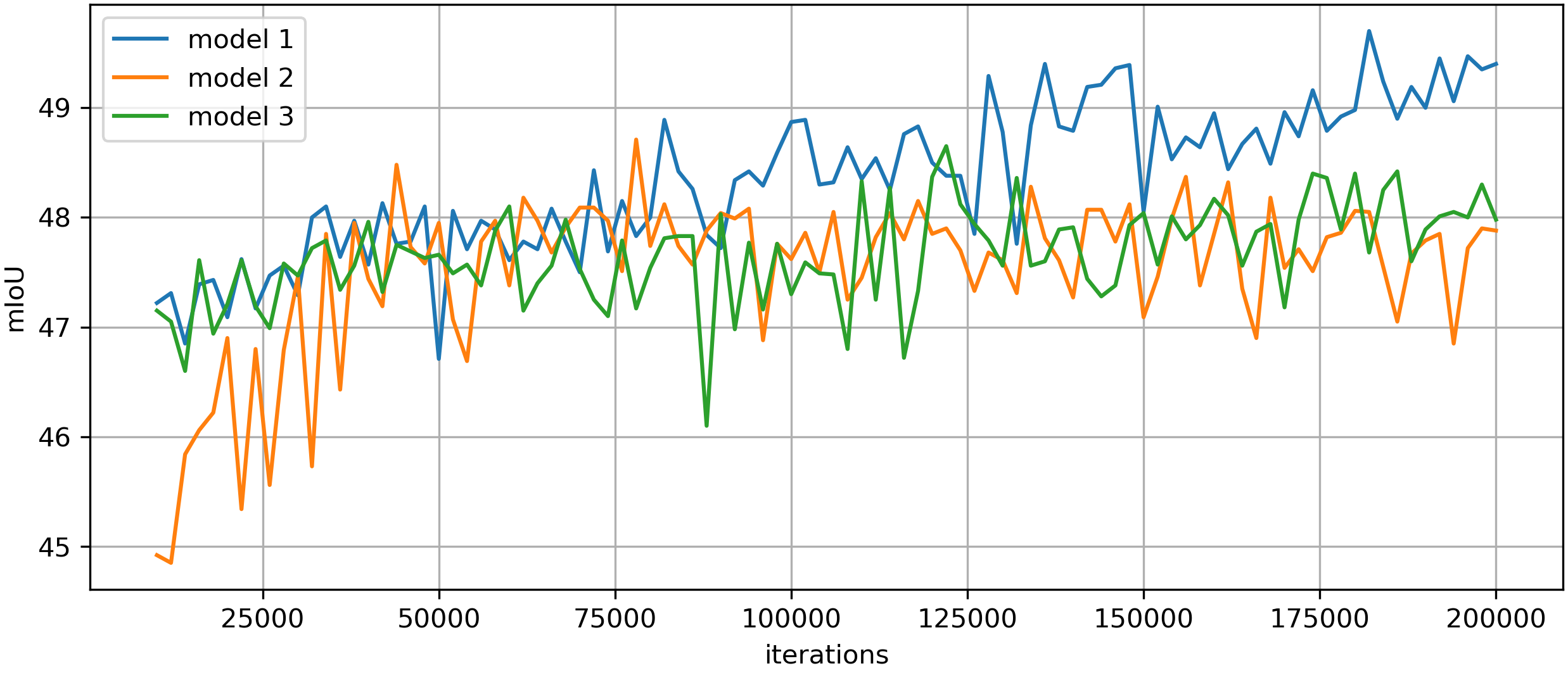}
    \caption{The ablation study on mIoU vs. iterations, model 1 - `Ours', model 2 - `with Adversarial', model 3 - `only SSL'.}
    \label{fig:ablation}
    \vspace{-5mm}
\end{figure}

\begin{figure*}[t]
    \centering
    \includegraphics[width = 1.0\linewidth]{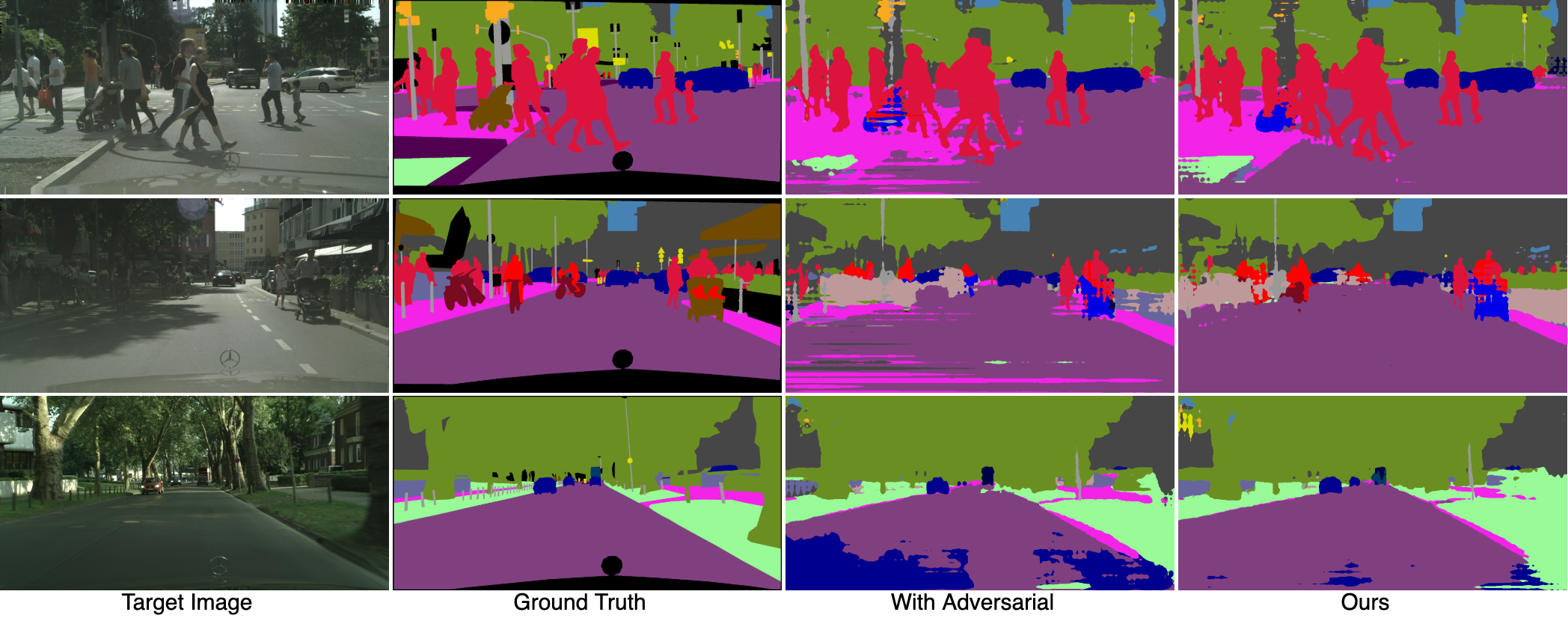}
    \caption{Qualitative Comparison. From left to right, it shows the target image, ground truth, prediction outputs of 'with Adversarial' and 'Ours'. 'Ours' shows much clear prediction outputs than 'with Adversarial'.}
    \label{fig:qualitative}
    \vspace{-3mm}
\end{figure*}

\subsection{Ablation Study}
\label{ablation_study}
In this section, we deeply investigate the contributions of our method via ablation study. 
By removing each building block of our work, we show its adequacy.
In Tab.~\ref{table:ablation} `w/o Dictionary' refers to training without the source feature dictionary thus the cosine similarity matrix is computed only between the current iteration's source features and target features. `w/o Class-wise Split' means without using the split process $\mathcal{A}$ and $\mathcal{B}$, thus it also does not use the dictionary and the cosine similarity matrix is computed between the current iteration's unsplit source and target features as a whole. `with Adversarial' is to use the adversarial adaptation loss along with our $\mathcal{L}_{total}$, (\ref{step2}). `only SSL' refers to training only with the segmentation loss, $\mathcal{L}_{seg}$. `SSL with Adversarial' is a model trained by $\mathcal{L}_{seg}$ with the adversarial adaptation loss.
As can be seen in the table, mIoU decreases when each contribution of our work is removed. The gap between `Ours' and `only SSL' shows the effectiveness of our cosine similarity loss. What is interesting is the results of `with Adversarial': the mIoU rather decreases which means the adversarial adaptation loss disturbs the training. The gap between `only SSL' and `SSL with Adversarial' also supports this. For SYNTHIA\textrightarrow Cityscapes, adversarial adaptation leads to marginal performance gain but for GTA5\textrightarrow Cityscapes, the result is rather worsen. 
This can be more clearly observed in Fig.~\ref{fig:ablation}. It shows the plots of mIoU on validation set at every 2,000 iteration during training. Model 1, 2 and 3 refer to `Ours', `with Adversarial' and `only SSL' models (DeepLabV2) trained on GTA5\textrightarrow Cityscapes respectively. The plots show a clear gap between `Ours' and other two models. The mIoU of model 2 increases fast until certain iteration but stops gaining any additional performance since then. Moreover, its plot fluctuates excessively which shows the instability of adversarial training. The clear gap between model 1 and model 2 shows the validity of our cosine similarity loss. Also, `Ours' constantly moves upwards as the iteration goes on, showing the possibility of further improvement in further iterations. 

\begin{table}[t]
		\begin{center}
		    \resizebox{0.75\linewidth}{!}{
			\begin{tabular}{l|l|ccc}
				\toprule
				Task & Arch. & $\mathcal{T}_{cos}$ & \textit{dict-size} & $\lambda_{cos}$ \\ 
				\midrule
				{\multirow{2}{*}{GTA5\textrightarrow CS}}
				 & R & 0.6 & 2500 & 0.01 \\
				 & V & 0.4 & 2000 & 0.001 \\
				\midrule
				{\multirow{2}{*}{SYNTHIA\textrightarrow CS}}
				& R & 0.4 & 2500 & 0.01 \\
				& V & 0.2 & 2000 & 0.001 \\
				\bottomrule
			\end{tabular}}
		\end{center}
		\vspace{-2mm}
		\caption{Hyper-parameters used for our method according to the task and the architecture.}
		\label{table:hyper-parameter}
		\vspace{-2mm}
\end{table}

\begin{table}[t]
		\begin{center}
		    \resizebox{0.9\linewidth}{!}{
			\begin{tabular}{l|ccccc}
				\toprule
				$\mathcal{T}_{cos}$ & 0.5 & 0.55 & 0.6 & 0.65 & 0.7\\
				mIoU & 48.92\% & 49.15\% & \bf49.7\% & 49.06\% & 49.47\% \\
				\midrule
				\textit{dict-size} & 1000 & 1500 & 2000 & 2500 & 3000 \\ 
				mIoU & 49.07\% & 49.51\% & 49.26\% & \bf49.7\% & 49.13\% \\
				\bottomrule
			\end{tabular}}
		\end{center}
		\vspace{-2mm}
		\caption{Hyper-parameters analysis. $\textit{dict-size}$ is set to 2500 while analyzing $\mathcal{T}_{cos}$ and $\mathcal{T}_{cos}$ is set to 0.6 while analyzing $\textit{dict-size}$.}
		\label{table:hyper-parameter_analysis}
		\vspace{-5mm}
\end{table}

\subsection{Hyper-parameter Analysis}
\label{hyper-params}
There are three hyper-parameters used in our cosine similarity loss which are $\mathcal{T}_{cos}$, $\textit{dict-size}$ and $\lambda_{cos}$. $\mathcal{T}_{cos}$ determines the amount of target features to be maximized. If it is set too low, almost every target features would be selected and possibly be maximized with even dissimilar source features. If it is set too high, not enough target features would be selected thus the cosine similarity loss would have no effect. $\textit{dict-size}$ decides how many source feature vectors are to be stored in the dictionary for each class. The larger its size, the older the features that can be stored. $\lambda_{cos}$ is simply used to balance the loss. $\mathcal{T}_{cos}$ and $\textit{dict-size}$ are important hyper-parameters of our cosine similarity loss, hence the values are different by architecture, dataset and pseudo labels, but we find that the hyper-parameters shown in Tab.~\ref{table:hyper-parameter} work in general. Tab.~\ref{table:hyper-parameter_analysis} shows our hyper-parameter analysis on GTA5\textrightarrow Cityscapes using DeepLabV2.

\subsection{Qualitative Comparison}
Fig.~\ref{fig:qualitative} shows qualitative comparison results. The first two columns are target images and the corresponding ground truth labels while the last two columns are qualitative results of ``with Adversarial" and ``Ours". As can be seen from the figure, when the adversarial adaptation loss is additionally applied, the prediction outputs are more noisy and not clear. It incorrectly classifies the road as some other classes. `Ours' has much smoother and clearer prediction outputs. Furthermore, our method performs better at recognizing distant objects, for example, people on the right side of the second row. We conjecture that this is due to our cosine similarity loss that tries to adapt target features to the specific source features that are most similar.

\section{Conclusion}
We propose maximizing cosine similarity between the source and the target domain at the feature level to tackle unsupervised domain adaptation. Our method measures the cosine similarity between a target feature and every source features spatially by classes and selectively maximizes the similarity only with the most semantically similar source features. 
We also propose source feature dictionary to maintain the latest source features which enables target features to be maximized with various source features and prevents the unmatching problem. We empirically show that our method can replace the unstable adversarial adaptation which is incapable of selectively adapting the target features to the most related source features.

\appendix
\counterwithin{figure}{section}
\counterwithin{table}{section}
\section{How to generate the pseudo-labels}

We basically follow the process of generating pseudo-labels proposed by \cite{li2019bidirectional}.
We assume a segmentation network $\mathcal{G}$ is already trained. As mentioned in the main paper, a segmentation network generates a prediction output $\mathcal{G}(x^t)=P^t \in \mathbb{R}^{H \times W \times C}$. By following the equation below, we generate a pseudo-label $\hat{y}^t$ that corresponds to an input image $x^t$. 
\begin{equation}
    \begin{split}
    &(C^{t}_{max}, P^{t}_{max}) = (\argmax_{c\in [C]} P^{t(c)}, \max_{c\in [C]} P^{t(c)}) \\
    &\hat{y}^{t} = \mathbbm{1}_{[ P^{t}_{max} > \tau^{C^{t}_{max}}]} \odot C^{t}_{max} \in \mathbb{R}^{H \times W}
    \end{split}
    \label{eq:labelpseudo-supple}
    \vspace{-2mm}
\end{equation}
We argmax along class dimension and filter only pixels whose prediction confidence exceed a class-specific confidence threshold $\tau^c$. $\tau^c$ is the confidence threshold of class $c$. $\tau^c$ is set by the confidence score of top 50\% of each class. We infer all the images in the training set into the network to obtain the prediction output for each image. Then, for each class, we collect all the prediction pixels that are classified as the class and add the confidence score of each pixel to a list. We sort the list in a descending order and choose the median value of the list as the $\tau^c$ for the class. If the median value is higher than 0.9, we set $\tau^c$ as 0.9. Therefore, each class has its own $\tau^c$ and it can be different by classes.

\section{Figure of process $\mathcal{A}$ and $\mathcal{B}$}

\begin{figure*}[t]
    \centering
    \includegraphics[width = 0.9\linewidth]{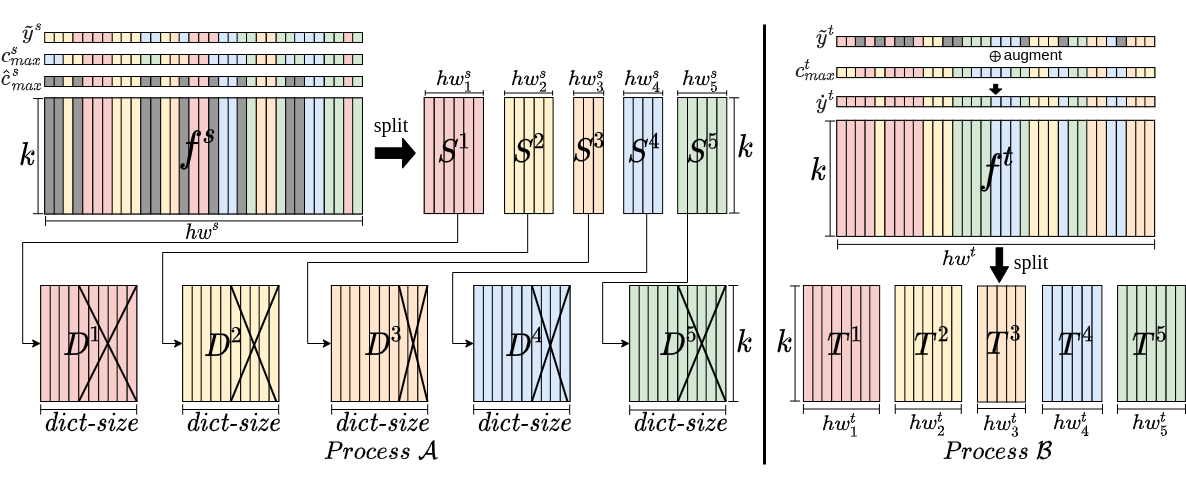}
    \caption{Grey shaded parts in process $\mathcal{A}$ are incorrectly classified source prediction outputs and features, while in $\mathcal{B}$, they are ignore symbols of the pseudo-label. Note that $\tilde{y}^t_j$ is augmented by replacing the ignore symbols with the predicted classes of the target output, $c^t_{max}$, generating $\dot{y}^t$.}
    \label{fig:split}
    \vspace{-4mm}
\end{figure*}

Fig.~\ref{fig:split} illustrates our process $\mathcal{A}$ and $\mathcal{B}$. The illustration of process $\mathcal{B}$ corresponds to the case when the pseudo-label is available. Process $\mathcal{A}$ first selects features that are correctly classified using $\hat{c}^s_{max}$. $\hat{c}^s_{max}$ is the correctly classified prediction output classes of $c^s_{max}$ via the ground truth label $\tilde{y}^s$. The grey shaded parts of $\hat{c}^s_{max}$ and $f^s$ are incorrectly classified prediction outputs and source features respectively. Then, we split the source features $f^s$ into classes according to $\hat{c}^s_{max}$. The split source features are enqueued to the dictionary class by class. 

Process $\mathcal{B}$ first augments the pseudo-label $\tilde{y}^t$ with the target prediction output. As it can be seen in the figure, a pseudo-label $\tilde{y}^t$ has ignore symbols where the confidence score of the separate trained network is lower than the confidence threshold $\tau^c$. We augment $\tilde{y}^t$ by replacing those ignore symbols with the prediction output classes of the current training network, $c^t_{max}$. $\dot{y}^t$ is the augmented pseudo-label and it is basically a copy of $\tilde{y}^t$ but has the values of $c^t_{max}$ where $\tilde{y}^t$ has ignore symbols. Then, we split the target feature map $f^t$ class-wise according to $\dot{y}^t$.

\section{Formulation of Adversarial Adaptation}
Along with the segmentation loss, usually an adversarial adaptation loss is adopted to make the distribution of $P_t$ closer to $P_s$ by fooling a discriminator network $\mathcal{D}$. It tries to maximize the probability of target predictions being considered as source predictions by $\mathcal{D}$.
\begin{equation}
    \mathcal{L}_{adv}(x^t) = -\sum_{h,w}\log(\mathcal{D}(P^t)^{(h,w)})
    \label{adv}
\end{equation}
On the other hand, $\mathcal{D}$ is trained to correctly distinguish the originating domain of the segmentation output.
\begin{equation}
\begin{split}
    \mathcal{L}_D(x^s,x^t) = -\sum_{h,w}(\log(1-\mathcal{D}(P^t)^{(h,w)})\\
    +\log(\mathcal{D}(P^s)^{(h,w)}))
    \label{discriminator}
\end{split}
\end{equation}
The discriminator network is a fully convolutional network which consists of 5 convolutional layers with $4\times4$ kernels and a stride of 2, the channel sizes are set as \{64, 128, 256, 512, 1\} for each layer respectively. The first four layers are followed by leaky ReLU \cite{agarap2018deep} parameterized by 0.2. We use ADAM optimizer with learning a rate of $1 \times 10^{-4}$ for DeepLabV2 based on ResNet101 and $1\times10^{-6}$ for FCN-8s based on VGG16. The momentums are set as 0.9 and 0.99. The loss balance parameter $\lambda_{adv}$ is set as 0.001 and 0.0001 for DeepLabV2 and FCN-8s respectively.
However, in our method, we do not employ any adversarial loss since we find it ineffective for self-supervised learning and moreover it disrupts the training when combined with our proposed cosine similarity loss.

\section{Tweaked method for FCN}
FCN-8s has a different architecture from DeepLabV2. The main difference is that it does not use the bilinear interpolation but instead uses skip combining to fuse the outputs from shallow layers and a transposed convolutional layer for upsampling. It combines the outputs from three different layers which are fc7, pool4 and pool3. Therefore, our cosine similarity loss is applied to these three different layers. We name the three different feature maps from the three layers as $\{f^s_l\}^3_{l=1}$. Since FCN-8s does not use the bilinear interpolation but rather directly produces the prediction output, we have to downsample the prediction output $P^s = \mathcal{G}(x^s) \in \mathbb{R}^{H \times W \times C}$ into three different spatial sizes of the feature maps $\{f^s_l\}^3_{l=1}$. We first argmax $P^s$ along the class dimension and obtain predicted class information.

\begin{equation}
    C^{s}_{max} = \argmax_{c\in [C]} P^{s(c)} \in \mathbb{R}^{H \times W}
\end{equation}
Then we resize $C^{s}_{max}$ into the three spatial sizes of $\{f^s_l\}^3_{l=1}$ via nearest interpolation. $c^s_{max} = I_{nearest}(C^s_{max})$
The three resized class outputs are $\{c^s_{max}{}_l\}^3_{l=1}$ which have the same spatial sizes as $\{f^s_l\}^3_{l=1}$ respectively. We also resize the ground-truth label $y^s$ to the spatial size of each $\{c^s_{max}{}_l\}^3_{l=1}$ thus generate three resized ground truth labels $\{\tilde{y}^s_l\}^3_{l=1}$. We use $\{c^s_{max}{}_l\}^3_{l=1}$ and $\{\tilde{y}^s_l\}^3_{l=1}$ to split the source feature maps analogous to (\ref{eq:source-split}) of the main paper.
\begin{equation} 
\begin{split}
    \hat{c}^s_{max}{}_l &= \mathbbm{1}_{[c^s_{max}{}_l = \tilde{y}^s_l]} \odot c^s_{max}{}_l \\
    S^c_l &= \mathbbm{1}_{[\hat{c}^s_{max}{}_l = c]} \otimes f^s_l.
\end{split}
\end{equation}
$S^c_l$ refers to the split source features from layer $l$ that is correctly classified as class $c$.
There are three different dictionaries each corresponding to each layer, $\{D_l\}^3_{l=1}$. 
Each $S^c_l$ is enqueued to $D^c_l$.

This process is analogously applied to the target feature map as well. 
We resize the target prediction output $P^t$ to the spatial size of the three target feature maps $\{f^t_l\}^3_{l=1}$, generating $\{c^t_{max}{}_l\}^3_{l=1}$. 
The pseudo-label $\hat{y}^t$ is also resized to the spatial size of each $\{c^t_{max}{}_l\}^3_{l=1}$ as $\{\tilde{y}^t_l\}^3_{l=1}$.
\begin{equation}
\begin{split}
    \dot{y}^t_l = augment(\tilde{y}^t_l, {c^t_{max}}_l) \\
    T^c_l = \mathbbm{1}_{[\dot{y}^t_l=c]} \otimes f^t_l.
\end{split}
\end{equation}
$T^c_l$ refers to split target features from layer $l$ that is classified as class $c$. 
\begin{equation}
\begin{split}
    \mathcal{M}^c_l = \frac{{T^c_l}^\mathsf{T} \boldsymbol{\cdot} D^c_l}{\norm{T^c_l}^\mathsf{T}_2 \boldsymbol{\cdot} \norm{D^c_l}_2} \\
    \hat{\mathcal{M}}^c_l = \mathbbm{1}_{[\mathcal{M}^c_l>\mathcal{T}_{cos}{}_l]} \odot \mathcal{M}^c_l.
\end{split}
\end{equation}
The cosine matrix is computed for each class of each layer. $\mathcal{M}^c_l$ is the cosine matrix of class $c$ between source features stored in $D^c_l$ and split target features from layer $l$ classified as class $c$. We only select elements of $\mathcal{M}^c_l$ that exceed $\mathcal{T}_{cos}{}_l$. $\mathcal{T}_{cos}$ is defined differently for each layer $l$. A higher threshold is defined for a shallower layer since the shallow layer possesses more global information without the detailed information while features from a higher layer posses high-level abstraction that is more detailed. Therefore, we want to set a higher threshold for a shallow layer in order to maximize the similarity of the target features with more meaningful and similar source features. We simply add 0.1 to the baseline threshold as the layer goes shallower, for example, if we set 0.5 as the baseline threshold, the $\mathcal{T}_{cos}{}_l$ for fc7, pool4 and pool3 layers are set as 0.5, 0.6, 0.7.
The final cosine similarity loss is averaged over the three layers.
\begin{equation}
\begin{split}
    \mathcal{L}_{cos}(x^t) =\frac{1}{3 \cdot C} \sum_{l=1}^3\sum_{c=1}^C\norm{\hat{\mathcal{M}}^c_l - \textbf{1}}^1_1.
\end{split}
\end{equation}

\section{More qualitative results}
\begin{figure*}[t]
    \centering
    \includegraphics[width = 0.95\linewidth]{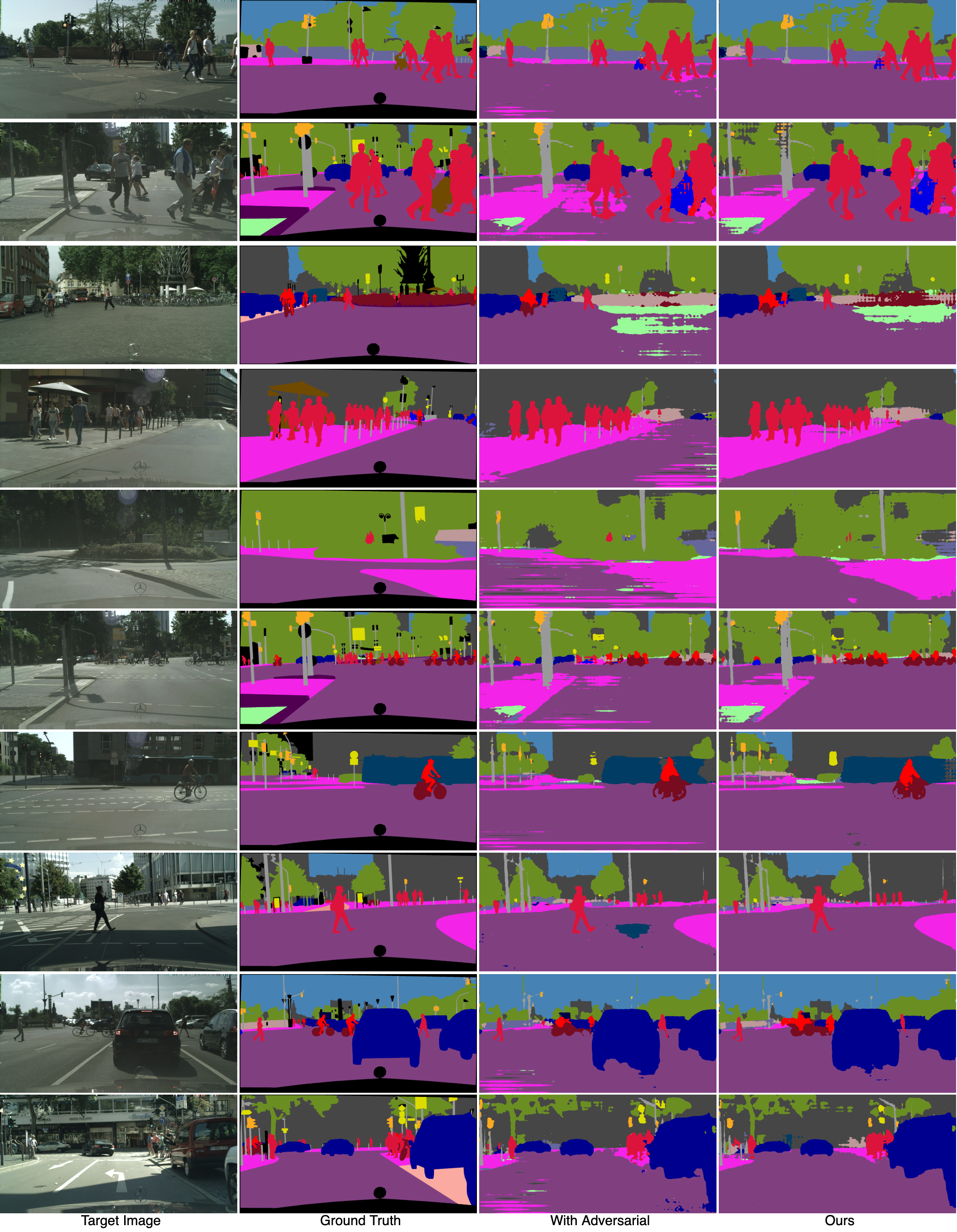}
    \caption{More qualitative results.}
    \label{fig:more-qualitative-results}
    \vspace{-4mm}
\end{figure*}
Fig.~\ref{fig:more-qualitative-results} shows more qualitative comparison results between ``with Adversarial" and ``Ours".

\clearpage
\newpage
{\small
\bibliographystyle{ieee_fullname}
\bibliography{egbib}
}

\end{document}